\pdfoutput=1
\documentclass[11pt]{article}

\usepackage{acl}

\usepackage{times}
\usepackage{latexsym}

\usepackage[T1]{fontenc}
\usepackage[utf8]{inputenc}

\usepackage{microtype}
\usepackage{multirow}
\usepackage{graphicx}
\usepackage[utf8]{inputenc}
\usepackage[arabic,farsi,english]{babel}
\usepackage{soul}
\usepackage{multicol}
\usepackage{multirow}
\usepackage{comment}

\title{MACRONYM: A Large-Scale Dataset for Multilingual and Multi-Domain Acronym Extraction}

\author{
Amir Pouran Ben Veyseh\textsuperscript{\rm 1}, Nicole Meister\textsuperscript{\rm 2}, \\
\textbf{Seunghyun Yoon\textsuperscript{\rm 3}, Rajiv Jain\textsuperscript{\rm 3}, Franck Dernoncourt\textsuperscript{\rm 3}}, \textbf{Thien Huu Nguyen\textsuperscript{\rm 1}}\\
\textsuperscript{\rm 1}Department of Computer and Information Science, \\University of Oregon, Eugene, OR, USA\\
\textsuperscript{\rm 2}Department of Electrical and Computer Engineering, \\Princeton University, Princeton, NJ, USA\\
\textsuperscript{\rm 3}Adobe Research, San Jose, CA, USA\\
{\tt \{apouranb,thien\}@cs.uoregon.edu} \\
{\tt nmeister@princeton.edu} \\
{\tt \{syoon,rajijain,franck.dernoncourt\}@adobe.com} \\
}

\begin{document}
\maketitle
\begin{abstract}

%including information extraction and machine translation

Acronym extraction is the task of identifying acronyms and their expanded forms in texts that is necessary for various NLP applications. Despite major progress for this task in recent years, one limitation of existing AE research is that they are limited to the English language and certain domains (i.e., scientific and biomedical). As such, challenges of AE in other languages and domains is mainly unexplored. Lacking annotated datasets in multiple languages and domains has been a major issue to hinder research in this area. To address this limitation, we propose a new dataset for multilingual multi-domain AE. Specifically, 27,200 sentences in 6 typologically different languages and 2 domains, i.e., Legal and Scientific, is manually annotated for AE. Our extensive experiments on the proposed dataset show that AE in different languages and different learning settings has unique challenges, emphasizing the necessity of further research on multilingual and multi-domain AE.

%We also conduct extensive analysis on the proposed dataset

%Acronym understanding is the task of identifying acronyms and their expanded forms in texts. It consists of extraction of acronyms and also their disambiguation. This is necessary for various NLP applications including question answering, information extraction and machine translation. Despite the major progress on this task in the recent years, one of the limitations of the existing works is that they are limited to the English language and certain domains (e.g., scientific or biomedicine). As such, challenges of acronym understanding in other languages and domains is mainly unexplored. To address this limitation, in this work, we propose a multilingual multi-domain acronym extraction and disambiguation dataset. Specifically, 27,200 documents in 6 typologically different languages and 2 domains, i.e., Legal and Scientific, have been manually annotated. We conduct extensive analysis on the proposed dataset. Our analyses show that AE in different languages has unique challenges, emphasizing the necessity of further research on multilingual AE systems.

\end{abstract}

\section{Introduction}

%One of the methods to reduce reading/writing time and facilitate communication involves using acronyms in lieu long phrases. Acronyms are short forms of longer phrases that are often constructed using a few letters selected from the long phrases. Due to their functionality, acronyms are common in many languages and domains. For instance, 73\% of abstracts of scientific papers contain at least an acronym \cite{barnett2020meta}. As such, in a text processing application, e.g., question answering, information extraction or machine translation, it is necessary to correctly identify the acronyms and their meanings. In the literature, acronym understanding is defined by two sub-tasks: (1) Acronym Extraction (AE) whose goal is to recognize the acronyms and their definitions in text. For instance, in the sentence ``\textit{They will meet in the conference of the World Trade Organization (WTO)}", an AE system identifies ``\textit{WTO}" and ``\textit{World Trade Organization}" as the acronym and long-form, respectively; (2) Acronym Disambiguation (AD) whose goal is to predict the correct meaning of an acronym given in a context. For instance, in the sentence ``\textit{The report is drafted from the input of related NGOs}", the acronym ``\textit{NGO}" should be disambiguated to ``\textit{Non-Governmental Organization}". 

%One of the methods to reduce reading/writing time and facilitate communication involves using acronyms in lieu long phrases. 

Acronyms are short forms of longer phrases that are often constructed using a few letters selected from the long phrases. Due to their functionality, acronyms are common in many languages and domains. For instance, 73\% of abstracts of scientific papers contain at least an acronym \cite{barnett2020meta}. As such, in text processing applications, e.g., question answering and machine translation, it is necessary to correctly identify the acronyms and their meanings. Toward this goal, our work focus on the task of Acronym Extraction (AE), aiming to recognize acronyms and their definitions/long forms in text. For instance, in the sentence ``\textit{They will meet in the conference of the World Trade Organization (WTO)}", an AE system should identify ``\textit{WTO}'' and ``\textit{World Trade Organization}" as the acronym and long form, respectively.

Despite all progress in recent years, prior works on AE are mainly limited to specific domains and languages. Specifically, biomedical and scientific texts in English have been the main focus in prior works. However, recognition of acronyms in other languages and domains is also important and might involve challenges not reflected in English biomedical/scientific texts. For instance, many existing AE methods for English employ uppercase letters to identify acronyms \cite{veyseh2020acronym}. However, in non-case sensitive languages, e.g., Arabic or Persian, uppercase letter concept does not exist, thus causing a failure of existing AE systems. Moreover, in each domain or language, different styles might be exerted to shorten a longer phrase to produce acronyms. For instance, initial letters of the words in the phrases are commonly used to form acronyms in scientific English; however, in legal English or Danish documents, the use of initial letters for acronym detection is less effective (see Section \ref{sec:exp}). As such, it is desirable to study AE in more diverse domains and languages to better support multi-domain and multilingual applications. 

Unfortunately, to the best of our knowledge, there is no existing dataset for multilingual and multi-domain AE, thus impeding research effort in this area. To this end, our work addresses this issue by introducing a new manually labeled dataset for AE. In particular, based on two different domains of scientific and legal texts, our dataset annotates AE data for sentences in six different languages: English, Danish, Spanish, French, Persian, and Vietnamese. As such, legal texts, Danish, Spanish, French, Persian, and Vietnamese are not explored for AE in prior work. In addition, our dataset is large-scale, providing 27,200 annotated sentences for AE to support advanced model development (e.g., with data-hungry deep learning models).

%Furthermore, based on the annotated AE data, we construct a multilingual and multi-domain dataset for AD, offering 1,832 documents in English, Spanish and French for AD task. 

%in this work, we propose a manually labeled dataset to address this shortcoming. In particular, for two domains of Scientific and Legal text, we collect 27,200 text excerpts in English, Danish, Spanish, French, Persian, and Vietnamese. Specifically, for the legal domain, we employ the United Nations Parallel Corpus and Europarl to collect data in English, Danish, Spanish and French. For the domain of Scientific documents, we download the ACL anthology and also available MS and PhD thesis from the internet to collect data in English, Persian and Vietnamese. All documents are manually annotated by native speakers of each language.  Furthermore, we employ the labeled data to construct a multilingual and multi-domain dataset for acronym disambiguation. Specifically, 1,832 documents in English, Spanish and French are provided for AD task.

%and explore promising directions for future work

%, using the prepared AE datasets, we conduct extensive experiments to study the limitations of the current state-of-the-art methods. 

Finally, we conduct extensive experiments to understand the challenges of the AE task in the created dataset. Our experiments show that the AE task in our dataset presents significant challenges for existing models in different domains and languages. This is even more pronounced in the cross-lingual and cross-domain transfer settings where existing models perform poorly on our AE dataset.  As such, more research effort is needed to address the challenges of acronym understanding in different settings. We will publicly release the dataset to foster research in this area.  

%Using the prepared AE and AD dataset, we conduct extensive experiments to study the limitations of the current state-of-the-art methods and to explore the promising directions for future work. Specifically, our analysis shows that AE and AD are challenging tasks in all languages, and more importantly, existing systems perform poorly in cross-lingual and cross-domain settings. As such, more research is needed to address the challenges of acronym understanding in these settings. We publicly release the data and models to foster research in this direction. 

\section{Data Annotation}
\label{sec:annotation}

%In this work, we provide manually labeled dataset for multilingual and multi-domain acronym extraction (called MACRONYM) in 6 languages and 2 domains. This section describes our data collection and annotation processes. 

%For more details, see the Appendix \ref{}. 

\textbf{Data Collection}: We collect data in two domains of legal and scientific documents for AE annotation. For each domain, documents in different languages are required. As such, for the legal domain, we employ the United Nations Parallel Corpus (UNPC) \cite{ziemski2016united} and the Europarl corpus \cite{koehn2005europarl}. The UNPC corpus contains official records in 6 languages while the Europarl corpus consists of the proceedings of the European Parliament in European languages. To accommodate our annotation budget and diversify the resulting dataset, we choose documents from four languages in the two corpora (i.e., English, French, and Spanish in UNPC, and Danish in Europarl) for our AE annotation. In addition, for the scientific domain, we employ the publicly available papers and M.S./Ph.D. theses in the field of computer science for AE annotation. Specifically, we collect the papers published in the ACL anthology of natural language processing research for English. Also, for typologically different languages, we crawl public computer science thesis in Persian and Vietnamese.

Following \cite{veyseh2020acronym}, we split the selected documents into sentences that will be annotated separately by annotators. In addition, to optimize the annotation cost with greater numbers of acronyms, we apply the same procedure in \cite{veyseh2020acronym} to filter out sentences that has low chance to contain acronyms or long forms. In particular, the procedure only retains sentences that involve at least one acronym candidate (i.e., a word with more than a half of characters as capital letters) and a sub-sequence of words to match the acronym candidate (i.e., concatenating the initials of the words can form the candidate) \cite{veyseh2020acronym}. Here, we only apply this procedure for English, French, Spanish, and Danish as our Persian and Vietnamese data is small and the sentence filtering procedure will leave less sentences for annotation. Finally, given the retained sentences for each language, we randomly sample a subset of sentences for manual AE annotation. The numbers of annotated sentences are presented in Table \ref{tab:stats}.

%in each language and domain

\noindent \textbf{Annotation Process}: To annotate the sampled sentences, we recruit native speakers in each language from the crowd-sourcing platform \url{upwork.com} with freelancer annotators across the globe. For each language, we select annotator candidates who have experience in related annotation projects and an approval rate of more than 95\% (provided by Upwork). The annotator candidates are trained with guidelines and examples for AE in their language. In our annotation guideline, acronyms are required to be single words (including abbreviations). Also, for a sentence in a language, we only annotate long forms that are in the same language as the sentence's. Afterward, for each language, we retain two candidates who pass and achieve highest results in our designed test for AE as our official annotators. Next, the two annotators in each language independently perform AE annotation for the sampled sentences of that language. Finally, the two annotators will discuss to resolve any disagreement in the annotation, thus producing a final version of our MACRONYM dataset.

\newcommand{\rota}[2]{\parbox[t]{4mm}{\multirow{#1}{*}{\rotatebox[origin=c]{90}{#2}}}}

%\begin{table}[]
%    \centering
%    \resizebox{.28\textwidth}{!}{
%    \begin{tabular}{l|c|c}
%        Language & \multicolumn{2}{c}{Domain} \\ \cline{2-3}
%        & Legal & Scientific \\ \hline
%        English & 0.824 & 0.811 \\
%        Spanish & 0.810 & - \\
%        French & 0.823 & - \\
%        Danish & 0.801 & - \\
%        Persian & - & 0.782 \\
%        Vietnamese & - & 0.791
%    \end{tabular}
%    }
%    \caption{IAA scores using Krippendorff’s alpha \cite{krippendorff2011computing} with the MASI distance metric \cite{passonneau2006measuring} for initial independent annotations in each domain and language of MACRONYM).}
%    \label{tab:IAA}
%\end{table}

\begin{table}[ht]
    \centering
    \resizebox{.46\textwidth}{!}{
    \begin{tabular}{c|l|c|c|c|c}
        \multicolumn{2}{c|}{Domain} & IAA & Size & \# Unique & \# Unique \\ %& Avg \# Acronyms  \\
        \multicolumn{2}{c|}{\& Language} & & & Acronyms & Long-forms \\ \hline %& per Paragraph \\ \hline
        \rota{4}{Legal}
        & English & 0.824 & 4,000 & 3,688 & 3,037 \\ % & 2.68 \\
        & Spanish & 0.810 & 6,400 & 4,059 & 4,437 \\ % & 2.18 \\
        & French & 0.823 & 8,000 & 5,638 & 5,728 \\ % & 2.78 \\
        & Danish & 0.810 & 3,000 & 907 & 923 \\ \hline % & 2.03 \\ \hline
        \rota{4}{Scientific}
        & English & 0.811 & 4,000 & 3,604 & 4,260 \\ % & 1.93 \\
        & Persian & 0.782 & 1,000 & 641 & 203 \\ % & 1.83 \\
        & Vietnamese & 0.791 & 800 & 270 & 61 \\ % & 1.05 \\
        %& & & & \\
    \end{tabular}
    }
    %\caption{Statistics of the proposed AE dataset MACRONYM. Size refers to the number of annotated paragraphs. The first half is for the legal domain and the second half is for the scientific domain.}
    \caption{\small Statistics of MACRONYM. IAA scores use Krippendorff’s alpha with MASI distance based on initial independent annotations. Size refers to the number of annotated sentences.}
    \label{tab:stats}
\end{table}

To study the challenges of AE in each language, following \cite{veyseh2020acronym}, we compute the inter-annotator agreement (IAA) scores using Krippendorff’s alpha \cite{krippendorff2011computing} with the MASI distance metric \cite{passonneau2006measuring} for the initial independent annotations of the two annotators, i.e., before resolving the conflicts. Table \ref{tab:stats} shows the IAA scores for each language. Overall, we find that the IAA scores are high for all considered languages and domains, thus demonstrating the quality of our annotated dataset. Among several factors, a major scenario of annotation disagreement occurs in Persian or Vietnamese when a sentence contains a long form term that is translated from an original English term. However, the acronym for this long form in the Persian or Vietnamese sentence is still formed via the initials of the words in the English term. As such, some annotators consider this English-based acronym as an acronym in the Persian or Vietnamese sentence while other annotators simply ignore it in the annotation. For instance, in the Persian sentence:

\begin{small}
``\FR{به شرح زیر است} (ANOP) \FR{عملیات پیشرفته شبکه}"\footnote{English translation: ``\textit{Advanced network operations (ANOP) include the followings}''},
\end{small}

the acronym ``\textit{ANOP}'' is expressed in English letters but its long form, i.e., ``{\small \FR{عملیات پیشرفته شبکه}}"\footnote{English translation: ``\textit{advanced network operations}"}, is presented in Persian. In the resolving, we have decided to annotate any acronym that is formed using characters in the six languages in our dataset.

%Among the six languages, there are more disagreement between the annotators in Persian and Vietnamese languages than the others. A major scenario of annotation disagreement in these two languages occurs when a sentence in Persian or Vietnamese contains a long form term that is translated from an original English term. However, the acronym for this long form in the Persian or Vietnamese sentence is still formed via the initials of the words in the English term. As such, some annotators 

%This table shows that while the IAA is high for all languages and domains, there are more disagreement between the annotators in Persian and Vietnamese languages than the others. We found that the use of symbols and also translation of the long-forms are the main two sources of disagreement in these two languages. For instance, in the sentence:

%\begin{small}
%``\FR{نشان دهنده مقاومت الکتریکی است} R \FR{در جدول زیر}"\footnote{English: In the following table R represents electrical resistance},
%\end{small}

%the symbol ``\textit{R}" should not be annotated as the acronym for the phrase ``{\small \FR{مقاومت الکتریکی}}"\footnote{English: electrical resistance}. 

%\subsection{Data Analysis}

\noindent \textbf{Data Analysis}: We show the main statistics of MACRONYM in Table \ref{tab:stats}. This table shows that the density of acronyms in texts varies across different languages. On average, English sentences tend to involve more acronyms than other languages in both legal and scientific domain while Danish and Vietnamese sentences contain least acronyms in the legal and scientific domain respectively. Comparing English texts in the legal and scientific domains, we find that the ratio between the numbers of unique long-forms and acronyms is greater in the scientific domain, thus implying the higher ambiguity of acronyms in scientific documents. Finally, we note that the number of unique acronyms exceeds the number of unique long forms in Persian and Vietnamese as we do not apply the sentence filtering procedure in the data collection, thus allowing many sentences with only acronyms and no associated long forms to be annotated in the data.

\begin{table*}[ht]
    \centering
    \resizebox{.82\textwidth}{!}{
    \begin{tabular}{l|l|c|c|c|c|c|c|c|c|c}
        \multicolumn{2}{l|}{Domain} & \multicolumn{3}{c|}{Mono-Lingual} & \multicolumn{2}{c|}{Mono-Lingual} & \multicolumn{2}{c|}{Cross-Lingual} & \multicolumn{2}{c}{Cross-Lingual} \\ 
         \multicolumn{2}{l|}{\& Language} & \multicolumn{3}{c|}{Mono-Domain} & \multicolumn{2}{c|}{Cross-Domain} & \multicolumn{2}{c|}{Mono-Domain} & \multicolumn{2}{c}{ Cross-Domain} \\ \cline{3-11}
         \multicolumn{2}{l|}{} &  \multicolumn{1}{c|}{Rule-Based} & \multicolumn{1}{c|}{mBERT} & \multicolumn{1}{c|}{XLMR} & \multicolumn{1}{c|}{mBERT} & \multicolumn{1}{c|}{XLMR} & \multicolumn{1}{c|}{mBERT} & \multicolumn{1}{c|}{XLMR} & \multicolumn{1}{c|}{mBERT} & \multicolumn{1}{c}{XLMR} \\ \hline
        % & F1 & F1 & F1 & F1 & F1 & F1 & F1 & F1 & F1 \\ \hline
        \rota{4}{Legal}
        & English & 16.55 & 61.66 & 62.07 & 54.92 & 56.88 & - & - & - & -  \\
        & Spanish & 10.82 & 51.43 & 55.41 & - & - & 38.88 & 40.13 & 35.48 & 36.92 \\
        & French & 10.05 & 58.77 & 61.14 & - & - & 48.82 & 50.70 & 44.21 & 46.83 \\
        & Danish & 8.78 & 50.05 & 48.38 & - & - & 40.71 & 42.94 & 38.18 & 41.95 \\ \hline
        \rota{4}{Scientific}
        & English & 20.72 & 60.51 & 59.00 & 56.71 & 59.88 & - & - & - & - \\
        & Persian & 60.59 & 62.41 & 63.10 & - & - & 49.13 & 50.21 & 42.95 & 43.72 \\
        & Vietnamese & 53.44 & 58.71 & 59.13 & - & - & 50.72 & 51.44 & 48.32 & 50.17 \\
        %& & & & & & & & &   & 
    \end{tabular}
    }
    \caption{\small Model performance (F1 scores) in different settings. The performance in each row is evaluated on the test data for the corresponding pair of language and domain. {\bf Mono-Lingual Cross-Domain}: trained on English data of one domain and tested on English data of the other domain; {\bf Cross-Lingual Settings}: trained on English data of one domain and tested on the other language data of the same domain (if Mono-Domain) or the other domain (if Cross-Domain).}
    
    %{\bf Cross-Lingual Mono-Domain}: trained on English data of one domain and tested on the other language data of the same domain; {\bf Cross-Lingual Cross-Domain}: trained on English data of one domain and tested on the other language data of the other domain.}
    
    %Cross-lingual settings involve training models on English data and testing them on other languages according to appropriate domain settings.}
    \label{tab:AE_results}
\end{table*}

\section{Experiments}
\label{sec:exp}
This section studies the challenges of the multilingual and multi-domain AE task in MACRONYM. In particular, for each pair of available languages and domains (we have 7 pairs in total), we first prepare the data by randomly splitting the corresponding set of annotated sentences into separate training/development/test portions with the ratios of 80/10/10 (respectively). Afterward, we report the performance of the representative AE models on the test set for each possible pair of languages and domains under different learning settings. 

%In this section we empirically study the challenges in multilingual AE and AD. First, for each language, we randomly split the paragraphs into train/development/test sets with the ratio 80/10/10. Next, we report the performance of the state-of-the-art models on the test set of each language and domain. For this analysis, we conduct experiments using the following three baselines in four different settings. 

\textbf{AE Models}: We examine the performance of three representative state-of-the-art (SOTA) models for AE. First, we employ the rule-based system for AE proposed in \cite{veyseh2021maddog} (called {\bf Rule-Based}). This system serves as the current SOTA rule-based method for AE \cite{veyseh2021maddog}. In general, to detect acronyms, words with more than 60\% characters as uppercase letters are selected. To find long-forms, if a detected acronym is bounded between parentheses and the initial letters of preceding words can form the acronym, the system predicts the preceding words a long form. Second, motivated by prior work \cite{veyseh2020acronym,zhu2021atbert}, we solve AE as a sequence labeling problem using BIO tagging schema. In particular, following the current SOTA deep learning model for AE \cite{zhu2021atbert}, we employ a pre-trained BERT-based language model followed by a feed-forward network layer with softmax in the end to predict BIO-based label for each word in the sentence. To facilitate the learning on multiple languages, we explore two multilingual transformer-based language models, i.e., mBERT \cite{devlin2019bert} and XLMR \cite{conneau2020unsupervised}, leading to two models {\bf mBERT} and {\bf XLMR} for this approach.

%In particular, we leverage the multilingual transformer-based language model mBERT \cite{devlin2019bert} due to its demonstrated high performance for many multilingual learning tasks \cite{wu2019beto}; and (3) \textbf{XLMR+BiLSTM+CRF}: This model is similar to \textbf{mBERT}; however, we replace mBERT with the multilingual pre-trained model XLMR \cite{conneau2020unsupervised}.

%(2) \textbf{mBERT+BiLSTM+CRF}: This model follows the deep learning model in \cite{veyseh2020acronym} that formulates AE as a sequence labeling problem. In particular, starting with the representation vectors for the words in a sentence, this model employs a Bidirectional LSTM (BiLSTM) network followed by a Conditional Random Field (CRF) layer to predict the BIO-based label sequences for the sentence for AE. To facilitate the learning in multiple languages in MACRONYM, we employ the multilingual transformer-based language models mBERT \cite{devlin2019bert} to encode texts and obtain word representation vectors for different languages in this model. mBERT has been shown to produce state-of-the-art performance for different multilingual learning tasks \cite{wu2019beto}; and (3) \textbf{XLMR+BiLSTM+CRF}: This model is similar to \textbf{mBERT+BiLSTM+CRF}; however, we replace mBERT with the multilingual pre-trained model XLMR \cite{conneau2020unsupervised}.

%To this end, we fine-tune mBERT with a sequence labeling head; \textbf{XLMR}: Similar to the previous baseline with the difference that we employ XLMR model to encode the input. 

\textbf{Settings}: MACRONYM enables the evaluation of AE models on four different settings: (i) \textbf{Mono-Lingual Mono-Domain}: In this setting, training and test data of the models come from the same language and domain. As we have 7 possible pairs of languages and domains, this setting involves 7 different evaluations for each AE model; (ii) \textbf{Mono-Lingual Cross-Domain}: Training and test data for models belongs to the same languages, but different domains in this setting. In MACRONYM, this setting is only possible for English where AE models are trained on the legal domain but tested on the scientific domain and vice versa (i.e., two possible evaluations).; (iii) \textbf{Cross-Lingual Mono-Domain}: Assuming the same domain for training and test data, this setting trains models on English training data and evaluate them on test data of other languages. We thus have 3 and 2 possible evaluations for the legal and scientific domains respectively.; (iv) \textbf{Cross-Lingual Cross-Domain}: Training and test data for models originates from different languages and domains in this setting. As such, we also consider five evaluations in this setting. In the first two evaluations, models are trained on English data in the legal domain and evaluated on Persian and Vietnamese test data in the scientific domains. In contrast, for the other three evaluations, English data in the scientific domain is used for model training while Spanish, French, and Danish test data in the legal domain is used for evaluation. We fine-tune the hyper-parameters for the models using the development data for each pair of languages and domains. 
% The selected values are presented in Appendix \ref{app:repo} along with a reproducibility checklist.

%\textbf{Mono-Lingual Mono-Domain}: The model is trained on the language-specific train sets in the same domain and language as the test set; \textbf{Mono-Lingual Cross-Domain}: The model is trained on the English data in Legal or Scientific and is evaluated on English Scientific or Legal test sets, respectively; \textbf{Cross-Lingual Mono-Domain}: We use the English data to train the baseline. Then, we evaluate the model on the test set of the other languages. Note that, for Persian and Vietnamese, we train the model on English Scientific domain, while for the rest we use English Legal domain; \textbf{Cross-Lingual Cross-Domain}: Similar to prior setting, we use English data for training and test sets of the other languages for evaluation. However, the model trained on English Legal data is evaluated on Persian and Vietnamese test sets and the model trained on English Scientific data is evaluated on the remaining languages. 

\textbf{Results}: Table \ref{tab:AE_results} presents the performance of three AE models in four different settings. Note that as the {\bf Rule-Based} system does not require training, its performance in the mono-lingual and mono-domain setting can be applied to other settings. There are several observations from the table. First, the {\bf Rule-Based} system achieves decent performance for Persian and Vietnamese, but performs poorly for other languages. The main reason has to do with the dominance of acronyms over long forms in Persian and Vietnamese data (see Table \ref{tab:stats}). This is in contrast to other languages where acronyms and long forms are more balanced. As acronyms can be identified more easily with rules than long forms, the {\bf Rule-Based} system is more effective in the data with much more acronyms of Persian and Vietnamese. Second, in the legal domain where long forms are better presented, the performance of the models on English is significantly better than those for other languages, thus demonstrating the more challenging nature of non-English language for AE. Third, compared to deep learning models, the significant lower performance of the {\bf Rule-Based} model in thhe monolingual and mono-domain setting signifies the brittleness of human-designed rules for AE that necessitates learning models to improve the portability of models to different languages and domains. Fourth, across all learning models and language-domain pairs for test data, the lower performance in the cross-lingual mono-domain setting compared to its mono-lingual counterpart suggests the difference between languages that hinder cross-lingual transfer learning for AE. Fifth, the cross-domain performance also under-performs their monolingual counterpart for almost all learning models and language-domain pairs for testing, thus highlighting domain shifts as an important challenge for AE. Finally, across all the learning settings, the performance of the AE models is still far from being perfect in MACRONYM, thus presenting ample opportunities for future research in this area.

\section{Related Work}

%Acronym extraction and disambiguation are two established tasks that have been studied in prior works in the last decades. Early attempts employed rule-based methods rule-based methods \cite{park2001hybrid,wren2002heuristics,schwartz2002simple,adar2004sarad,nadeau2005supervised,ao2005alice,kirchhoff2016unsupervised} or feature engineering \cite{kuo2009bioadi,liu2017multi,wang2016clinical,li2018guess}. In recent years, deep learning is proved to be the SOTA method for both AE and AD \cite{veyseh2021maddog,wu2015clinical,antunes2017biomedical,charbonnier2018using,ciosici2019unsupervised,jaber2021participation,li2021systems}. Despite all progress in recent years, AE and AD in non-English languages is less explored. While some of the prior works propose non-English acronym glossaries \cite{pomares2020leveraging,menard2011classifier}, there are some limitations in them. First, the are limited to certain domain, e.g., biomedical domain. Second, they do not provide labeled data for AE task. To address these limitations, we propose the first large-scale multi-domain multi-lingual AE and AD dataset. 

%Acronym extraction is an established task that has been studied in prior works. 

%ao2005alice
%wang2016clinical

Early attempts for AE have employed rule-based methods \cite{park2001hybrid,wren2002heuristics,schwartz2002simple,adar2004sarad,nadeau2005supervised,kirchhoff2016unsupervised} or feature engineering models \cite{kuo2009bioadi,liu2017multi,li2018guess}. Recently, deep learning methods have delivered SOTA performance for AE \cite{veyseh2021maddog,wu2015clinical,antunes2017biomedical,charbonnier2018using,ciosici2019unsupervised,jaber2021participation,li2021systems}. Despite such progress, prior AE research and datasets have mainly focused on English biomedical and scientific texts, leaving non-English languages and other domains less explored. Here, we note that there exist some acronym glossaries for non-English languages \cite{pomares2020leveraging,menard2011classifier}. However, such resources do not annotate sentences/texts in multiple languages and domains for AE as we do in MACRONYM.

%is the first large-scale multilingual and multi-domain dataset that is manually annotated for AE.

%Lacking multilingual and multi-domain labeled datasets has been a major hindrance for this research direction.

%While some of the prior works propose non-English acronym glossaries \cite{pomares2020leveraging,menard2011classifier}, there are some limitations in them. First, the are limited to certain domain, e.g., biomedical domain. Second, they do not provide labeled data for AE task. To address these limitations, we propose the first large-scale multi-domain multi-lingual AE and AD dataset.

\section{Conclusion}
We present the first multilingual and multi-domain dataset for AE, involving annotation for 6 languages and 2 domains. Our experiments show that the proposed dataset presents significant challenges for AE methods in different learning settings and languages. In the future, we will expand the dataset to include more domains and languages for AE.

%, thus calling for more research effort in this area.

%We study the challenges of AE in the proposed dataset in various learning settings.

%In this work, we presented the first multilingual multi-domain acronym extraction dataset. The data is manually labeled and contains more than 27,000 samples in 6 different languages and 2 domains. Furthermore, we also present a dataset for acronym disambiguation task consisting of more than 27,000 samples in 4 languages and 2 domains. We study the challenges of the proposed dataset in various setting. Our analysis show that both AE and AD are challenging tasks in non-English languages. 

\bibliographystyle{acl_natbib}
\bibliography{acl_latex}

\clearpage

\appendix

\end{document}